\documentclass[runningheads]{llncs}
\usepackage{graphicx}
\usepackage{arydshln}
\usepackage{booktabs}
\usepackage{multirow}
\usepackage{multicol}
\usepackage{amsmath}
\usepackage{amssymb}
\RequirePackage{tikz}
\usepackage{pgfplots} 
\usepackage{amsmath}
\usepackage{hyperref} 
\usepackage{xurl} 

\usepackage{xcolor,colortbl}
\renewcommand{\paragraph}[1]{\noindent\textbf{#1}\quad}
\newcommand{\probP}{\text{I\kern-0.15em P}}

\renewcommand{\paragraph}[1]{\noindent\textbf{#1}\quad}
\definecolor{dgreen}{rgb}{0.0, 0.5, 0.0} 
\definecolor{aliceblue}{rgb}{0.94, 0.97, 1.0}
\definecolor{antiquewhite}{rgb}{0.98, 0.92, 0.84}
\definecolor{beaublue}{rgb}{0.9, 0.9, 0.98} %4
\definecolor{atomictangerine}{rgb}{0.88, 0.9, 0.8}%3

\definecolor{blond}{rgb}{0.9, 0.9, 0.98} %2

\newcommand{\correctpred}[1]{\textbf{\texttt{\textcolor{dgreen}{#1}}}}
\newcommand{\wrongpred}[1]{\textbf{\texttt{\textcolor{red}{#1}}}}

\newcommand{\convAA}[1]{\multicolumn{2}{l}{{\parbox[l]{0.45\textwidth} {\cellcolor{aliceblue}{#1}}}}}
\newcommand{\convAB}[1]{\multicolumn{2}{l}{{\parbox[l]{0.45\textwidth} {\cellcolor{antiquewhite}{#1}}}}}

\newcommand{\convBa}[1]{\multicolumn{3}{l}{{\parbox[l]{0.5\textwidth} {\cellcolor{aliceblue}{#1}}}}}
\newcommand{\convBb}[1]{\multicolumn{3}{l}{{\parbox[l]{0.58\textwidth} {\cellcolor{antiquewhite}{#1}}}}}
\newcommand{\convBc}[1]{\multicolumn{3}{l}{{\parbox[l]{0.58\textwidth} {\cellcolor{beaublue}{#1}}}}}

\begin{document}
%
% \title{Contribution Title\thanks{Supported by organization x.}}
\title{BiosERC: Integrating Biography Speakers Supported by LLMs for ERC Tasks}
\titlerunning{Integrating Biography Speakers Supported by LLMs for ERC Tasks}

\author{Jieying Xue  \and Minh-Phuong Nguyen  \and Blake Matheny   \and Le-Minh Nguyen} 

\institute{Japan Advanced Institute of Science and Technology\\
\email{\{xuejieying,phuongnm,matheny.blake,nguyenml\}@jaist.ac.jp} }
\maketitle              % typeset the header of the contribution
\begin{abstract}
% ###-------------rules------------------###
% The abstract should briefly summarize the contents of the paper in
% 15--250 words.
% ###-------------rules------------------###
In the Emotion Recognition in Conversation task, recent investigations have utilized attention mechanisms exploring relationships among utterances from intra- and inter-speakers for modeling emotional interaction between them. However, attributes such as speaker personality traits remain unexplored and present challenges in terms of their applicability to other tasks or compatibility with diverse model architectures. Therefore, this work introduces a novel framework named BiosERC, which investigates speaker characteristics in a conversation. By employing Large Language Models (LLMs), we extract the ``biographical information'' of the speaker within a conversation as supplementary knowledge injected into the model to classify emotional labels for each utterance. Our proposed method achieved state-of-the-art (SOTA) results on three famous benchmark datasets: IEMOCAP, MELD, and EmoryNLP, demonstrating the effectiveness and generalization of our model and showcasing its potential for adaptation to various conversation analysis tasks. Our source code is available at \href{https://github.com/yingjie7/BiosERC}{\url{https://github.com/yingjie7/BiosERC}}.

\end{abstract}
%
%
%

% ========================
\section{Introduction}
% ========================

% ############
% overview of erc task recently 
Emotion recognition in conversation (ERC) is a pivotal research topic that has garnered growing attention due to its extensive range of applications \cite{ghosal2019dialoguegcn,zhang2019modeling}.
% ,  mental health counseling \cite{althoff2016large}, and so on. 
% The objective of this task is to predict the emotional categories for each utterance given the conversation as context information.
In ERC tasks, the input text frequently consists of transcribed spoken dialogues from a speech recognition system, featuring colloquial or truncated statements that lack standardized grammar, thereby complicating emotional recognition in the dialogue.
Unlike the traditional non-conversation sentiment analysis task, ERC emphasizes some of the many factors that influence ERC tasks, including contextual and speaker-specific information \cite{ghosal2019dialoguegcn}.

Therefore, recent approaches have inclined toward encoding acoustic features \cite{hu-etal-2022-unimse_audio2,shi23e_interspeech} or contextual information \cite{li2020hitrans,liu-etal-2022-dialogueein,li2021past} to enrich utterance vector representation. 
On the other hand, numerous previous works have typically utilized GRU \cite{majumder2019dialoguernn,lee-lee-2022-compm,ijcai2022p562}, GNN \cite{poria2021recognizing,zhang2019modeling}, or self-attention network \cite{ghosal2019dialoguegcn,shen-etal-2021-directed,accwr} to encode richer speaker-specific information, including intra- and inter-speaker features. However, this latent information is predominantly learned from relationships among utterances. It poses challenges for validating its effectiveness and applying it to alternative tasks, and is problematic for other model architectures.
\begin{figure}[htbp]
    \centering
    \includegraphics[width=0.7\columnwidth,trim={0.cm 0.cm 0 0.cm},clip]{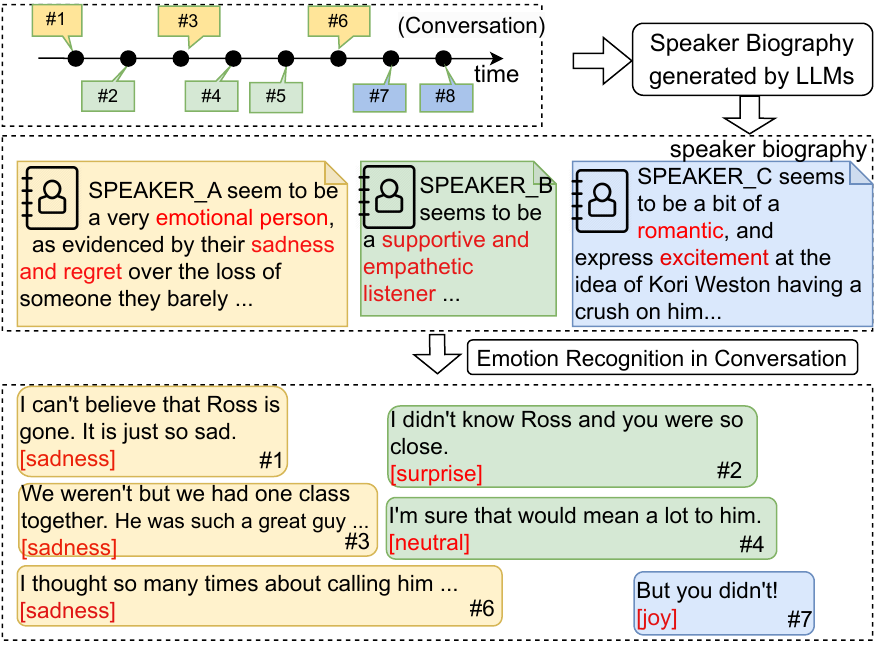}
    \caption{Overview of our BiosERC framework   
    % the red bar is "median values" average = 66.58 | cls=64.89 | lstm = 65.56 | mlp = 66.67
    }
    \label{fig:sen_ablation}
\end{figure}
% ####
% how important of speaker characteristic / speaker bio 
% However, the majority of prior research has predominantly concentrated on modeling individual speaker utterances or interactions among different speakers, with particular attention given to the intra- and inter-speaker aspects for the extraction of speaker-based information. Regrettably, limited emphasis has been placed on exploring speaker characteristics, which constitute critical and foundational elements of conversational information.
Additionally, speaker characteristics as a crucial and foundational feature in ERC tasks has not been comprehensively explored. We posit that within a dialogue, an individual's character can significantly influence their manner of emotional expression and habitual vocabulary selection, leading to varying emotions for the same statement even when articulated by different speakers. Comprehension of interlocutors' personality traits can thus facilitate accurately discerning their emotional inclinations within the discourse.
% Additionally, it is noteworthy that ERC tasks encounter challenging prominent phenomena known as ``emotional inertia'' and "emotional stimulation" \cite{liu-etal-2022-dialogueein}, significantly increasing the difficulty of sentiment recognition.
% Notably, speakers demonstrate two prominent phenomena known as "emotional inertia" and "emotional stimulation" during conversations \cite{liu-etal-2022-dialogueein}. This implies that the same speaker tends to maintain emotional consistency over brief time periods, resulting in similar or identical emotions, until an emotional stimulus intervenes. From another perspective, individuals who consistently employ sarcastic expressions may exhibit a greater complexity in the emotional content of their words. The speakers with fluctuating emotional states tend to have more extreme emotional tendencies. These variables collectively accentuate the substantial influence of speaker personality on the emotional dynamics within dialogues.

To tackle the aforementioned challenge, we propose BiosERC, a novel method designed to discover speakers' personality information to enhance ERC systems.
In contrast to previous methodologies relying on GRU \cite{lee-lee-2022-compm,ijcai2022p562} or speaker-based masked attention mechanisms \cite{ghosal2019dialoguegcn,shen-etal-2021-directed,accwr}  to capture emotional expression features of different speakers, BiosERC stands out by precisely extracting individual personalities of speakers within dialogues (Figure~\ref{fig:sen_ablation}). This uniqueness empowers the model to intricately comprehend character traits and encapsulate events of emotional transitions occurring within the characters. Moreover, our mechanism for extracting speaker characteristics is explicit and more amenable to verification and adaptation for application to various conversation analysis tasks. 

Specifically, BiosERC utilizes LLMs with a prompting technique \cite{wei2022chain,touvron2023llama} to extract descriptions of interlocutor features as supplementary knowledge, which are then injected into the emotion recognition process within conversations.
As shown in Figure~\ref{fig:sen_ablation}, this conversation involves three distinct speakers, each presenting unique perspectives and exhibiting markedly different emotional states.
The speaker description facilitates the model's thorough understanding of each speaker's role within a conversation. Particularly, SPEAKER A is experiencing \textit{sadness and regret} (as mentioned in the speaker description), resulting in expressions predominantly filled with sadness. SPEAKER B appears to be a supportive and \textit{empathetic listener}, with limited involvement in the conversation, and reacts through SPEAKER A's utterances. Meanwhile, SPEAKER C responds with \textit{excitement} upon hearing their conversation. 
Intuitively, the integration of biographical data plays an important role in enriching the emotional background of each speaker in conversations, and holds the potential for more precise and comprehensive emotional recognition, especially in complex dialogues.
% Furthermore, our approach excels in multi-party conversations by discerning intra-speaker relations through the extraction of individual speakers' personality traits. 
% Concurrently, it identifies the inter-speaker dynamics, elucidating the connections and influences among utterances of different speakers. Of notable significance, our speaker description not only extracts personality characteristics but also discerns other pertinent information such as the relational dynamics among multiple speakers.

We carry out experiments on three benchmark datasets, including IEMOCAP, MELD, and EmoryNLP. The experimental results demonstrate that our method achieves SOTA performance, which indicates the effectiveness of our proposed model. Furthermore, our proposed mechanism, which uses a prompting technique for LLMs to extract the speakers' biographical information, shows the potential to adapt to various conversation-level tasks such as opinion analysis, recommendation, and others.

\section{Related Work \label{sec:related_work}}
% % ========================
\paragraph{Emotion Recognition in Conversation.}
    % Emotion Recognition in Conversation (ERC) aims to discern the emotions of each utterance in the conversation. 
    In contrast to the conventional non-conversation sentiment analysis task, ERC demands a greater reliance on contextual and speaker-specific information for its support.
    % \cite{hazarika-etal-2018-conversational,ghosal-etal-2020-cosmic,liu-etal-2022-dialogueein,li-etal-2017-dailydialog}.
    For the purpose of modeling the conversational context, numerous studies employ Recurrent Neural Networks (RNNs) \cite{ghosal-etal-2020-cosmic,li-etal-2017-dailydialog} or Graph Convolution Network (GCN)   \cite{ghosal2019dialoguegcn,lee-choi-2021-graph} to explore the hidden relationships between utterances. 
    % This is applied due to its capability to model context-sensitivity through the training of graph structures.
    Moreover, the incorporation of contextual information and external knowledge into utterance vector representations has been notably achieved in recent works \cite{shen-etal-2021-directed,liu-etal-2022-dialogueein,ghosal-etal-2020-cosmic,Li_Zhu_Mao_Cambria_2023} through the utilization of self-attention mechanisms and pre-trained Language Models (LM) \cite{devlin2018bert,liu2019roberta}. In the recent success of LLMs on various NLP tasks, InstructERC \cite{lei2023instructerc} is proposed to utilize the instruction prompting technique and fine-tune the LLM model for ERC tasks. MKFM framework \cite{tu-etal-2023-empirical} proposed the utilization of diverse supplementary knowledge information  (e.g., emotional cause, topics) by ChatGPT service to inject into a graph-based model. In comparison, our work focuses on modeling speaker characteristics, a fundamental information which can be extracted by open-source LLMs (e.g, LLama-2). In addition, we also prove our proposed mechanism worked effectively when fine-tuning on both popular architectures: \textit{BERT} and \textit{transformer-based decoder-only LLM}.     %
    
    \paragraph{Speaker-based ERC.}
    Because of the significant impact of speakers on ERC, researchers have placed emphasis on speaker modeling. DialogueRNN \cite{majumder2019dialoguernn} and COSMIC \cite{ghosal-etal-2020-cosmic} leverage Gated Recurrent Units (GRU) for the modeling of speaker-specific semantic context. 
    % Since  graph-based methods are proficient in modeling context and speaker-sensitivity, 
    Some researches \cite{poria2021recognizing,zhang2019modeling} treat conversations as graphs while incorporating prior speaker information as distinct relationships between utterances, or considers speakers as nodes within the graph. 
    % DialogXL \cite{shen2021dialogxl} obtains richer speaker context information by using speaker-aware self-attention to capture the intra- and inter-speaker information.S 
    HiTrans \cite{li2020hitrans} exploits an auxiliary task to classify whether two utterances belong to the same speaker to make the model speaker-sensitive. S+PAGE \cite{liang-etal-2022-page} employs a two-stream conversation Transformer architecture to extract both self and inter-speaker contextual features. However, the majority of prior research has predominantly concentrated on modeling individual speaker utterances or interactions among different speakers, with particular attention given to the intra- and inter-speaker aspects for the extraction of speaker-based information \cite{liu-etal-2022-dialogueein,accwr}. Regrettably, limited emphasis has been placed on exploring speaker characteristics, which constitute critical and foundational elements of conversational information. Therefore, we propose a novel method named BiosERC, which employs external tools to extract speaker characteristics and inject them into the process of emotion recognition within conversations.    
    
% ========================    
\section{Methodology\label{sec:method}}
% ========================    
    This section introduces our baseline model architecture for the ERC task, which utilizes intra- and inter-speaker information following current SOTA methods \cite{liang-etal-2022-page,liu-etal-2022-dialogueein,ijcai2022p562}, and our proposed method BiosERC, which incorporates the biography of the speakers into an ERC model.  Formally, we define a conversation as: $\mathcal{C} = \{u_i\}_{ 0\leq i < |\mathcal{C}|}$, where each individual utterance $u_i$ is articulated by speaker $p(u_i) \in \mathcal{S}$, with $\mathcal{S}  = \{ s_j\}_{0\leq j < |\mathcal{S}|}$ representing the set of speakers in the conversation. Here, $p$ denotes a mapping function that associates utterances with their respective speakers. 
    % The objective of the task is to determine the emotion expressed in each utterance ($u_i$) from a predefined set of emotions.
    % , which may include categories like happiness, sadness, and others.

    \subsection{Intra-inter ERC (baseline)} 
        Based on recent SOTA methods in the ERC task \cite{liang-etal-2022-page,ijcai2022p562,accwr}, we implement our baseline model consisting of three principal components: utterance vector representation, context modeling, and an emotion classification layer.
        % \begin{itemize}
        
        \paragraph{Utterance Vector Representation.} To enrich meaning representations, we follow an approach that mixes the surrounding utterances within a fixed-window size \cite{emoberta,lee-lee-2022-compm,Li_Zhu_Mao_Cambria_2023,accwr}. Particularly, to encode a sentence $u_i$, the text input is combined by surrounding the utterance according to the following template: ``\textit{\textsc{[cls]}, $u_{i-w}$,  \texttt{..},  \textless/s\textgreater, $u_i$, \textless/s\textgreater, \texttt{...} $u_{i+w}$}'', where $w$ is the local contextual window size hyperparameter. 
            The utterance vector is computed by aggregating the respective word vectors following \cite{accwr}:
            \begin{align}
                 h^{cls}, h^{words} &= \mathrm{RoBERTa}( [u_{i-w},.. , u_{i+w} ])\label{eq:utt_word_accwr} \\
                 h^{utt} &=  [ \mathrm{tanh} ( \mathrm{average}(h^{\mathrm{words\,of\,u_i}} ) \cdot W^u )]_{ 0\leq i < |\mathcal{C}|}\label{eq:u_i}
            \end{align}
            where $h^{\mathrm{words\,of\,u_i}}$ denotes the word vectors selected from $ h^{words}$ at the positions of the utterance $u_i$; $h^{utt}$ is all utterance vectors in a conversation; and $W^*$ refers to learnable weights. 
            % The hidden states of all utterances in a conversation are then forwarded to the next component to learn the contextual information. 
            
        \paragraph{Context Modeling.} Utterance vectors are integrated contextual information of whole conversation by attention mechanism: 
        % \textit{global-context}, and \textit{intra-, inter-speaker} \cite{liu-etal-2022-dialogueein}. 
            \begin{align}
                \mathrm{Attn} (q, k, v, M) &= \mathrm{softmax} ( \frac{ q \cdot k^{\intercal} }{ \sqrt{d_t}}  + M) \cdot  v   \\
                {q}_t, {k}_t, {v}_t &= {h^{utt}}{W}_t^q , {h^{utt}}{W}_t^k , {h^{utt}}{W}_t^v  \label{eq:qkv_tf}\\
                {head}_t &= \mathrm{Attn}({q}_t, {k}_t, {v}_t, M) \label{eq:head_linear} \\
                {h}_{\mathrm{MultiHead}} &= \mathrm{concat}([{head}_{t |  0 <  t \leq H}]) {W}^o \label{eq:h_attn}
            \end{align}
            where $H$ is the number of heads in the MultiHead attention layer; ${q}_t, {k}_t, {v}_t$ are utterance vectors in various semantic space   (dimension size $d_t$).
            % $M$ is a masked matrix to control the dependencies among utterances.
            % \begin{align}
            % \end{align} 
            In detail, following \cite{liu-etal-2022-dialogueein,accwr}, we construct the relation matrices ($M$) for modeling relationship among utterances, where $M_{ik} = 0$ if $u_i$ and $u_k$ should have interaction, $M_{ik} = -\infty$ if otherwise. For the baseline model, we implement three different relationships: \textit{global context} (all utterance pairs are connected), \textit{intra-speaker} (only utterance pairs of the same speaker are connected), and \textit{inter-speaker} (only utterance pairs of the different speaker are connected). Consequently, we acquire three new hidden states (from Equation~\ref{eq:h_attn}) $h^{contxt}$, $h^{intra}$, $h^{inter}$ feed-forward to the Classification component.
            
            \paragraph{Classification.} This component aims to integrate all the hidden features of utterances to classify the emotion label. 
            \begin{align}
                h^{speaker}_i &=  {h^{intra}_i}{W}^a + {h^{inter}_i}{W}^r \label{eq:hspeaker}\\
                e^{o}_i  &=  {\mathrm{softmax} (h^{utt}_i}{W}^u + {h_i^{contxt}}{W}^g +  h^{speaker}_i  )  
            \end{align} 
            Then, the emotion vector ($e^{o}_i$) is used to compute the loss via \textit{Cross-Entropy} function and is trained based on the gold emotional label of the $i$-\textit{th} utterance. 
        % \end{itemize}

    \subsection{Bios ERC}
        In this section, we describe the process of generating the speaker's biography and present our BiosERC framework, leveraging two popular pre-trained LM as backbones: a BERT-based model  \cite{liu2019roberta} (e.g., RoBERTa) and a transformer-based decoder-only LLM model \cite{touvron2023llama} (e.g., Llama-2). Notably, we also introduce an effective mechanism using the biography of speakers incorporating fine-tuning a LLM-based \cite{radford2019language_gpt2} with the prompting technique. 
        
        \subsection{Biography of Speaker}
        In this part, we introduce a mechanism using the prompting technique for the LLMs to generate the description ($d_j$) for the respective speaker ($u_j$).  Given a conversation $\mathcal{C}$, the output of this step is the biography (description) of all speakers in a conversation $\mathcal{B} = \{d_j\}_{0\leq j < |\mathcal{S}|}$. 
        \begin{align}
            d_j = \mathrm{LLMs} (\mathrm{prompting}(\mathcal{C}, s_j))
        \end{align}
        \textit{LLMs} refers to large language models such as Llama2 \cite{touvron2023llama}, which can generalize a speaker's biography based on their conversation. The \textit{prompting} function is a template containing two conversation instances ($\mathcal{C}$) and speaker identification ($s_j$) to exploit the knowledge of the LLMs (Table~\ref{tab:prompting}). To avoid long plain text descriptions, we instruct the LLMs to limit the output by adding a \textit{``note''} concerning the length of the prompting template. Consequently, we obtain additional data about the persona of the speakers in each conversation ($\mathcal{B}$), which is utilized for speaker modeling in the subsequent step.
        \begin{table}[htbp]
        \centering
        \caption{Prompting template to extract the description of characteristics of the speaker from a conversation with LLMs.\label{tab:prompting}}
        \resizebox{.95\columnwidth}{!}{
        \begin{tabular}{|p{0.99\columnwidth}|}
            \hline
            Given this conversation between speakers: \\
            % "\\
            \texttt{\{conversation content $\mathcal{C}$\}}\\
            % "\\
            In overall above conversation, what do you think about the characteristics of speaker \texttt{\{speaker identification $ s_j$\}}? (Note: provide an answer within 250 words) \\ 
            \hline
        \end{tabular}
        }
        \end{table}
        
        \subsection{BERT-based BiosERC architecture\label{sec:bioserc_mlp}}
\begin{figure}[ht]
                \centering
                \includegraphics[width=0.7\columnwidth,trim={0.cm 0cm 0 0.cm},clip]{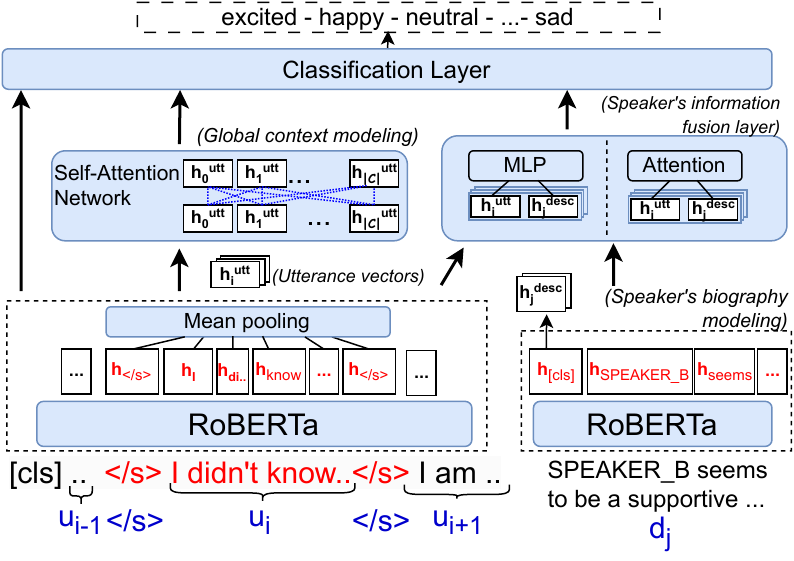}
                \caption{Overview of our BiosERC model architecture.\label{fig:overview_bioserc}}
            \end{figure}

            Firstly, we encode the speaker's description using a pre-trained language model to acquire hidden vector representation ($h^{desc}_j$).
            \begin{align}
                 h^{desc}_j &= \mathrm{RoBERTa}(d_j)[0]
            \end{align}
            where $j$ is the speaker index in the set of speakers in a conversation, $0\leq j < |\mathcal{S}|$. 
            % Then this vector is integrated with the corresponding utterance vector in the classification component. 
            Our proposed method, BiosERC, extends the baseline model and redefines the speaker's hidden vector representation ($h_i^{speaker}$ in Equation~\ref{eq:hspeaker}) (Figure~\ref{fig:overview_bioserc}). 
            % This is achieved with the following two mechanisms which model speaker persona information: a simple Multilayer Perceptron (MLP) or an attention layer. 
            This architecture is designed with a straightforward target that injects the personality information of each speaker into their corresponding utterances by a \textbf{m}ulti-\textbf{l}ayer \textbf{p}erceptron network. Then, the speaker information in Equation~\ref{eq:hspeaker} is replaced by:
            \begin{align}
                h_i^{speaker} &=  h^{desc}_{p(u_i)}  {W}^{desc} + b^{desc}
            \end{align}
            where $p(u_i)$ denotes the corresponding speaker of utterance $u_i$. Through this mechanism, all the utterances from the same speaker are shared in the unified speaker vector representation, while the weights are updated in the training process. Finally, the utterance vector is fused with the speaker vector which supports emotional classification.

            \paragraph{BiosERC - biography injected by attention mechanism.} We consider a variant of our BiosERC model,  which is engineered to dynamically incorporate the speaker's information into each utterance via the attention mechanism.  The relationship between the current utterance and all individual speakers is integrated to enrich the utterance vector representation. 
            \begin{align}
                h_i^{fusion} &=  h^{desc}_{p(u_i)}  {W}^{p} + h^{utt}_i\\
                h^{desc} &=  \{h^{desc}_j\}_{0\leq j < |\mathcal{S}|}\\
                h^{speaker}_i &=  \mathrm{Attn}(h_i^{fusion}, h^{desc}, h^{desc}, \mathbf{0})
            \end{align}
            We first compute a fusion vector ($h^{fusion}$) between the utterance and respective speaker description vectors. Then we collect all the speaker description vectors ($h^{desc}$) and use the attention mechanism to model the relationship between the utterance and all speakers in a conversation. Finally, the speaker features are embedded in this vector, $h^{speaker}_i$, and are replaced using Equation~\ref{eq:hspeaker} in the baseline system. 
        \subsection{LLM-based BiosERC + instruction fine-tuning  (ft LLM) \label{sec:bioserc_llm}}
        Since the robust natural language understanding capabilities of LLMs \cite{touvron2023llama}, we provide the speaker description as part of the text prompting input for the model (highlighted in blue in Table~\ref{tab:ft_prompting}) instead of modifying model architecture. We follow the instruction fine-tuning approach \cite{flant5_instruction}, with \textit{causal language modeling} objective to train an LLM to generate emotional label text (highlighted in red in Table~\ref{tab:ft_prompting}):
        \begin{align}
             x &=\textrm{prompting}(u_i, s_j, d_j, \mathcal{C}, e_i) \\ 
            \probP(x)&=\Pi_{z=1}^{|x|} \probP(x_z|x_0, x_1, ..., x_{z-1}) 
        \end{align}
        where $x, z$ is a sequence of tokens and the token's index in prompting input (Table~\ref{tab:ft_prompting}), respectively. 
        Additionally, we utilize LoRA \cite{hu2022lora}, a lightweight training technique, to reduce the number of trainable parameters. The instruction fine-tuned LLM learned the distribution of emotional labels given prompting input ($x$). During the inferring phase, the emotional label ($e_i$) which is omitted from prompting input, is left to be generated by the fine-tuned LLM.
        \begin{table}[htbp]
        \small
        \centering
        \caption{Prompting input template using speaker description and content of conversation  for fine-tuning LLMs.\label{tab:ft_prompting}}
        \resizebox{\columnwidth}{!}{
        \begin{tabular}{|p{1.15\columnwidth}|}
            \hline
            \textit{system}\\
            \#\#\# You are an expert at analyzing the emotion of utterances among speakers in a conversation. \\
            \textcolor{blue}{\#\#\# Given the characteristic of this speaker, \texttt{\{speaker name $s_j$\}}:}
            \textcolor{blue}{\texttt{\{speaker description $d_j$\}}}\\
            \#\#\# Given the following conversation as a context  
            \texttt{\{conversation $\mathcal{C}$\}}\\
            \textit{user}\\
            Based on the above conversation and characteristics of the speakers, which emotional label of \texttt{\{$s_j$\}} in the utterance \texttt{\{utterance $u_i$\}} ?\\
            \textit{assistant} \\
            \textcolor{red}{\texttt{\{emotional label of $u_i$ in text: $e_i$\}}}\\
            \hline
        \end{tabular}
        }
        \end{table}
    
% ========================
\section{Experimental Setting  \label{sec:experiment}  }
% ========================

    \paragraph{Datasets}
    We conducted evaluations on three ERC benchmark datasets in text-only version:  IEMOCAP \cite{busso2008iemocap}, involving daily conversations between pairs with ten different speakers; MELD \cite{poria2018meld}, derived from TV shows and featuring multiparty conversations; EmoryNLP \cite{zahiri2018emotion}, another multiparty daily dialogue dataset sourced from TV shows. The statistical information of these datasets is shown in Table~\ref{tab:stats_data}.  In accordance with prior works \cite{ghosal-etal-2020-cosmic,shen-etal-2021-directed}, we employed the Weighted-F1 score as the evaluation metric to maintain compatibility. 
    \begin{table}[!htbp]
    \small
        \centering
        \caption{
        Statistical information {on} all ERC datasets.      \label{tab:stats_data} }
        % \resizebox{\columnwidth}{!}{%
            \begin{tabular}{lcccccccc}
                \toprule  \multirow{2}{*}{\textbf{Dataset}}&\multicolumn{3}{c}{\textbf{\#dialogues\;\;\;\;\;}}& \multicolumn{3}{c}{\textbf{\#utterances\,\;\;\;}} & \multirow{2}{*}{\textbf{\#speaker}}   \\
                & train & dev &test &  train & dev &test \\
                \midrule
                IEMOCAP   &108& 12&31 & 5,163 & 647 &1,623 & 2.00  \\ 
                EmoryNLP  & 659  & 89 & 79 & 7,551 & 954  & 984 & 3.34 \\
                MELD  & 1,039 & 114 & 280 & 9,989 & 1,109 & 2,610&2.72 \\  \bottomrule
            \end{tabular}
            % } 
    \end{table} 
        
    \paragraph{Implementation Details}
    Since the recent successful applications and advancing capabilities of pre-trained LLMs  \cite{touvron2023llama,wei2022chain}, we leverage \texttt{LLama-2} model to procure personality descriptions for each participant in the conversation. Specifically, we verify the effectiveness of speaker description information on two aforementioned pre-trained LMs: the BERT-based model with \texttt{roberta-large} and the transformer-based decoder-only LLM model with \texttt{Llama-2-13b}.  
    The best model is determined based on the development set of each dataset and employed to evaluate the test set.
    For fine-tuning BERT-based BiosERC (section~\ref{sec:bioserc_mlp}), the hyper-parameters were selected as follows: the learning rate is selected from $\{1e^{-5}; 5e^{-6}\}$; the dropout value is $0.2$, and number epochs is $30$; and the local context window size ($w$) is chosen in $\{2, 4\}$;  we report the average scores obtained across 10 independent runs. For fine-tuning LLM-based BiosERC (section~\ref{sec:bioserc_llm}),  learning rate is selected from $\{2e^{-4}; 3e^{-4}\}$, number epochs is $3$;  we report the average scores obtained across 5 independent runs because of the computation cost. All the source code of this project is published at \texttt{MASKED\_LINK}.

% ========================
\section{Results and Analysis   \label{sec:result_analysis}}
    \subsection{Main results}
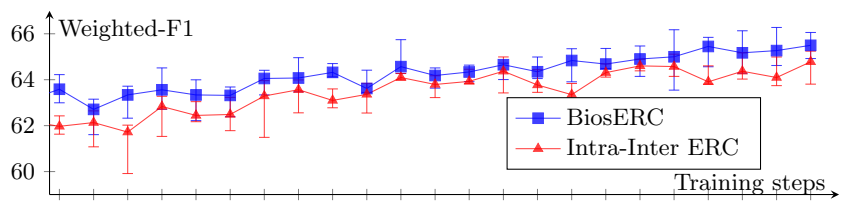
\begin{figure}[]
            \small
            \begin{tikzpicture}
                \begin{axis}[width=0.99\textwidth,height=0.33\textwidth,
                    axis lines=middle,
                    ymin=59, ymax=67,
                    xmax=8000, xmin=2000,
                    x label style={at={(current axis.right of origin)},anchor=south,above=4,left=1.9},
                    legend style ={at={(0.74,0.53)}, anchor=north, align=left},
                    xlabel= Training steps,
                    ylabel=Weighted-F1,
                    xticklabel style = {rotate=60,anchor=east},
                    xticklabel,
                    enlargelimits = false,
                    legend cell align={left}, 
                    xtick=data]
    
                    \addplot [opacity=0.75,  mark=square*, color=blue, error bars/.cd,  y dir=both, y explicit,] table [ x=Steps, y=bioserc_avg, col sep=space,
                    y error minus expr=\thisrow{bioserc_avg}-\thisrow{bioserc_min},
                    y error plus expr=\thisrow{bioserc_max}-\thisrow{bioserc_avg},
                        ]   {./learning_curver.tex};
                    \addlegendentry{ BiosERC}
                    \addplot [ opacity=0.75,mark=triangle*, color=red,  error bars/.cd, y dir=both, y explicit,] table [ x=Steps, y=baseline_avg, col sep=space,
                            y error minus expr=\thisrow{baseline_avg}-\thisrow{baseline_min},
                            y error plus expr=\thisrow{baseline_max}-\thisrow{baseline_avg},
                        ]   {./learning_curver.tex};
                    \addlegendentry{Intra-Inter ERC  } 
                \end{axis} 
            \end{tikzpicture}
            \caption{Performance comparison between our BERT-based BiosERC and the baseline model  (MELD dev set), illustrating the performance variability across 10 random runs. \label{fig:learning_process_dev}}
            % \endminipage
        \end{figure}

        Our approach demonstrated competitive performance compared to recent SOTA methods on three famous benchmark datasets (Table~\ref{tab:mainresult}) on both two architectures BERT-based and transformer-based decoder-only LLM model.
        % For this result, the \textit{BiosERC} architecture was used (section~\ref{sec:bioserc_mlp}) because this architecture showed better performance on the development set, which can be further analyzed in the next parts. 
        \begin{table}[]
            \centering
            \caption{
            Performance comparison between our proposed method and previous works on the test sets. Column \textit{\#T.Params.} refers to the number of trainable parameters. The notations $\ddagger$, $\dagger$ indicate the significant difference (t-test) with the baseline in levels $p < 0.01$ and $p<0.05$, separately. 
            % The values marked (*) mean that the evaluation metric is denotation match that different from others using sentence-level accuracy [checking] note in the review response.
            % This table contains two  parts: previous works and our results. 
            % The values below refer to our experimental results.  
            % The mark \textit{star} ($^*$) indicates that these works used external knowledge or additional data.   
            \label{tab:mainresult} }
            % \resizebox{0.4\textwidth}{!}{%
                \begin{tabular}{lrccc}
                    \toprule
                    \multirow{1}{*}{\textbf{Methods}} & \textbf{\#T.Params.} &\textbf{IEMOCAP}&    \textbf{EmoryNLP}  &    \textbf{MELD}     \\
                    % &  W-F1  &  W-F1  &   W-F1   \\
                    \midrule
                    % COSMIC  \cite{ghosal-etal-2020-cosmic}$^*$ & 65.28   & 38.11 & 65.21 \\ 
                    % CESTa \cite{wang-etal-2020-contextualized} &65.47 & - & 58.36  \\ 
                    HiTrans \cite{li-etal-2020-hitrans} && 64.50  & 36.75 & 61.94    \\
                    % SKAIG \cite{li-etal-2021-past-present} & &66.96 & 38.88 & 65.18   \\
                    DAG \cite{shen-etal-2021-directed} &&68.03 & 39.02 & 63.65  \\
                    DialogXL \cite{Shen_Chen_Quan_Xie_2021}&&  65.94 &  34.73&   62.14    \\
                    DialogueEIN \cite{liu-etal-2022-dialogueein} && 68.93 & 38.92  & 65.37   \\ 
                    % \hdashline
                    % SKIER \cite{Li_Zhu_Mao_Cambria_2023}$^*$ & - & \textbf{40.07}  & \textbf{67.39}   \\ 
                    % \qquad\textit{(Speaker-based methods)}\\
                     SGED + DAG-ERC \cite{ijcai2022p562} && 68.53 &40.24& 65.46\\
                     S+PAGE \cite{liang-etal-2022-page} &&68.93& 40.05 &64.67 \\
                    InstructERC  \cite{lei2023instructerc} \textit{+(ft LLM)} &&\textbf{71.39} & 41.39 &69.15\\\midrule
                    % \qquad\textit{(Our implementations)}\\
                    % \textsc{Baseline ERCß} &   &  &   \\ 
                    \textsc{Intra/inter} ERC (baseline) \cite{accwr} & $189 \times 10^6$  &   67.65 &  39.33 & 64.58  \\ 
                    \textsc{BiosERC}$_\textit{\texttt{BERT-based}}$  & $186 \times 10^6$ &  67.79 &{39.89}$^\dagger$ & 65.51$^\ddagger$ \\ 
                    \textsc{BiosERC}  \textit{+ft LLM$_\texttt{Llama-2-7b}$}  & $80 \times 10^6$ &  {69.02}  & {41.44} &{68.72} \\ 
                    \textsc{BiosERC}  \textit{+ft LLM$_\texttt{Llama-2-13b}$} & $125 \times 10^6$  &  {71.19}&\textbf{41.68}  & \textbf{69.83} \\ 
                    \bottomrule
                \end{tabular}
            % }
        \end{table} 
        % In particular, our model attains the highest performance on the MELD dataset and competes favorably with the best-performing model on the EmoryNLP and IEMOCAP datasets. In comparison to the baseline model, our model outperforms the baseline model across all datasets. More specifically, our model surpasses the baseline model by 0.14 on the IEMOCAP dataset, 0.56 on the EmoryNLP dataset, and 0.95 on the MELD dataset. 
        In comparison with the previous speaker-based methods (SGED + DAG-ERC \cite{ijcai2022p562},  S+PAGE \cite{liang-etal-2022-page} and DialogueEIN \cite{liu-etal-2022-dialogueein}) experimental results demonstrated the effectiveness of our proposed approach and further affirm that speaker modeling by speaker descriptions are superior to the information offered by intra- and inter-speaker contexts. In addition, our BiosERC model achieved significant differences with the baseline system on both the EmoryNLP and MELD datasets, which is shown clearly in Figure~\ref{fig:learning_process_dev}.  Because the MELD and EmoryNLP are multiparty conversation datasets (the average number of interlocutors are 2.72 and 3.34, respectively), the emotions are influenced more by different speaker personalities in a conversation than the IEMOCAP dataset. 
        
         In the previous method of fine-tuning an LLM,  InstructERC \cite{lei2023instructerc} considers speaker identifier as an auxiliary task, requiring two-stage training, which is more time-consuming than ours in the training process. Besides, our proposed method uses speaker descriptions generated by LLM in natural language, which can be easier incorporated with humans using our system for customization (e.g., customer support staff can directly provide or modify characteristics generated by LLM of their customers). Similar to BERT-based BiosERC, among three datasets, the LLM-based BiosERC shows the strengthens of multi-party datasets (more than two speakers in each conversation), EmoryNLP and MELD.  By fine-tuning an LLM, \texttt{Llama-2-13b}, the performance of our BiosERC increased by 1-4\% weighted F1 scores compared to BERT-based models and achieved new SOTA performance on EmoryNLP and MELD datasets.  Besides, since utilizing a lightweight training technique, LoRA \cite{hu2022lora}, the number of training parameters in LLM-based BiosERC was smaller than BERT-based BiosERC (only fine-tuned on two last layers) which proved the potential of LLM-based BiosERC in the real application. 
         
    \subsection{Ablation study  \label{subsec:ablation}}
        We conducted an ablation study to evaluate the effectiveness of integrating speaker biographies into the broader system encompassing various aspects.
        % : variants of speaker biographies generated by diverse LLMs and comparing with the baseline model. 
        % All ablation experiments were conducted on the MELD development set.
    
    \paragraph{BiosERC architecture.}  
        As shown in Table~\ref{tab:ablation}, it is apparent that our BERT-based BiosERC (row 3), which incorporates the speaker's descriptions, exhibits significant advantages in F1 score, outperforming the baseline system that relies on intra/inter- speaker relationships. Besides, by using the attention mechanism to encode the speaker's biography (row 2), BiosERC achieved high performance and it also clearly outperformed the baseline model. Moreover, in the setting of \textit{BiosERC  +fine-tuning LLM} (row 8), when removing the speaker description (the blue part in Table~\ref{tab:ft_prompting}) from the input prompting (row 6), the performance significantly decreased by 1.05 F1 score. By fine-tuning the different LLM models, \texttt{Llama-2-13b} and \texttt{Llama-2-7b}, the performance is slightly decreased with $0.52$ F1 score  (rows 7, 8).  These results proved the importance of the speaker's biography information and the efficacy of our proposed approach for speaker modeling. 
        
        % These results unequivocally demonstrate the robustness and effectiveness of our proposed approach.
        % outperforms the baseline system that relies on intra/inter-relationships through a self-attention mechanism. 
        % In Figure~\ref{fig:learning_process_dev}, the learning curve lines of our BiosERC (MLP) beat the baseline model on all checkpoints with significant margins on both F1 and loss scores. Compared with BiosERC (attention), although BiosERC (MLP) uses fewer learnable parameters, it is straightforward and provides useful information to the ERC system, avoiding overfitting problems in the learning process.  From the overall result of ten different runs, the performance distribution of BiosERC is higher than the baseline model.  
        % This superiority is evident in both the BiosERC models, whether using MLP or the attention mechanism. These results clearly 
        % \begin{figure}[ht]
        %     \centering
        %     \includegraphics[width=0.98\columnwidth]{images/method_ablation_boxplot.pdf}
        %     \caption{Ablation methods incorporating speaker biographies into the ERC task compared to baseline model.}
        %     \label{fig:intergrate method}
        % \end{figure} 
        \begin{table}[htbp]
            \centering
            \caption{
            Performance comparison among variants of BiosERC on the MELD development set.    
            \label{tab:ablation} }
            \resizebox{0.99\textwidth}{!}{%
                \begin{tabular}{lll}
                    \toprule
                    \multirow{1}{*}{\textbf{Methods}}&\textbf{LLMs extracting bio. }&   \textbf{Weighted-F1}      \\
                    % &  W-F1  &  W-F1  &   W-F1   \\
                    \midrule
                    1. \textsc{Intra/inter} ERC (baseline) &  - &  $66.08$$_{(-1.19)}$ \\ \cdashline{1-3}
                    2. \textsc{BiosERC} \textit{injecting bio. by attention}  & \texttt{Llama-2-chat-70b} & {$66.71^\dagger$}$_{(-0.56)}$ \\ 
                    3. \textsc{BiosERC}  &  \texttt{Llama-2-chat-70b} & \textbf{$\mathbf{67.27}^\ddagger$} \\ 
                    4. \textsc{BiosERC}   &  \texttt{Llama-2-chat-7b} & {$67.23^\ddagger$}$_{(-0.04)}$ \\ 
                    5. \textsc{BiosERC}   &  \texttt{vicuna-33b-v1.3} & {$66.96^\ddagger$}$_{(-0.32)}$ \\ \midrule
                    6. \textsc{BiosERC}  +ft LLM$_\texttt{Llama-2-13b}$ \textit{w/o speaker bio.} &  - & {$69.17_{(-1.05)}$} \\ \cdashline{1-3}
                    7. \textsc{BiosERC}  +ft LLM$_\texttt{Llama-2-7b}$  &  \texttt{Llama-2-chat-70b} & {$69.70_{(-0.52)}$} \\
                    8. \textsc{BiosERC}  +ft LLM$_\texttt{Llama-2-13b}$   &  \texttt{Llama-2-chat-70b} & {$\mathbf{70.22^\dagger}$} \\
                    \bottomrule
                \end{tabular}
            }
        \end{table} 
    
    % Regarding the window size of the local context (the hyper-parameter $w$ in Equation~\ref{eq:utt_word_accwr}), we conducted an ablation study ranging from a window size of 2 to 4, (as shown in Figure~\ref{fig: local context window}). The results indicate that a window size of 4 yields the best outcome. This suggests that the window size also affects the model's performance. In shorter conversations, window size has a relatively minor impact, but in longer conversations, a window size of 4 clearly outperforms the size of 2. Consequently, to conserve computational resources, we employed a window size of 4 in the IEMOCAP dataset and a window size of 2 in the MELD and EmoryNLP datasets.
    % - the cuver line showed on Figure~\ref{fig:learning_process_dev} \\
    % ======================== 
    % \begin{figure}[ht]
    %     \centering
    %     \includegraphics[width=0.98\columnwidth]{images/method_ablation_window.pdf}
    %     \caption{Performance comparison in change of local context window size in the utterance vector representation component.}
    %     \label{fig: local context window}
    % \end{figure}   

    \paragraph{Speaker biographies.}  
    We explored various currently popular LLMs for generating speaker biographies, including \texttt{LLama-2-chat-70b}, \texttt{Llama-2-chat-7b} \cite{touvron2023llama}, and \texttt{vicuna-33b-v1.3} \cite{Zheng2023JudgingLW}. Among these, \texttt{LLama-2-chat-70b} yielded the best outcomes. Based on observation, we found that the \texttt{Vicuna} model failed to provide speaker descriptions in some \textit{``extra difficult''} cases, such as when the conversation length is too short (e.g., less than three utterances) or when the specific speaker has extremely short utterances (e.g., Hmm). These solid improvements  worked on the diverse biographies generated by various LLMs underscore the versatility and effectiveness of extracting ``speaker biographies'', demonstrating that the LLMs framework can be highly beneficial for biography generation and assisting with ERC tasks. 
    % This substantiates the applicability of our proposed approach, which focuses on extracting pertinent features from a distinct perspective, and extends its potential utility to analogous research domains.

    \subsection{Conversation length}
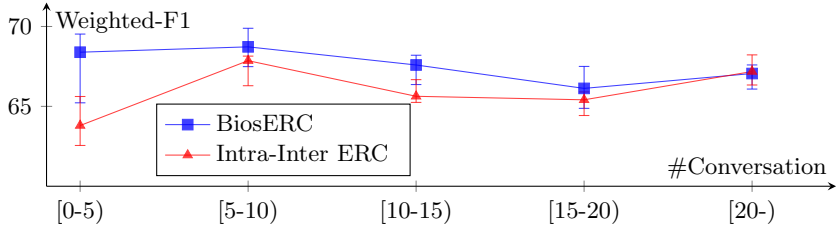
\begin{figure}[hpbt]
            \small
            \begin{tikzpicture}
                \begin{axis}[width=0.99\textwidth,height=0.33\textwidth,
                    axis lines=middle,
                    ymin=60, ymax=71.5,
                    xmax=5.5, xmin=0.8,
                    x label style={at={(current axis.right of origin)},anchor=south,above=7,left=0.2},
                    legend style ={at={(0.3,0.45)}, anchor=north, align=left},
                    xlabel= \#Conversation,
                    ylabel=Weighted-F1,
                    xticklabels from table={./length_analysis_processed.tex}{length},
                    enlargelimits = false,
                    legend cell align={left}, 
                    xtick=data]
    
                    \addplot [opacity=0.75,  mark=square*, color=blue, error bars/.cd,  y dir=both, y explicit,] table [ x=idx, y=bioserc_avg, col sep=space,
                    y error minus expr=\thisrow{bioserc_avg}-\thisrow{bioserc_min},
                    y error plus expr=\thisrow{bioserc_max}-\thisrow{bioserc_avg},
                        ]   {./length_analysis_processed.tex};
                    \addlegendentry{ BiosERC}
                    \addplot [ opacity=0.75,mark=triangle*, color=red,  error bars/.cd, y dir=both, y explicit,] table [ x=idx, y=baseline_avg, col sep=space,
                            y error minus expr=\thisrow{baseline_avg}-\thisrow{baseline_min},
                            y error plus expr=\thisrow{baseline_max}-\thisrow{baseline_avg},
                        ]   {./length_analysis_processed.tex};
                    \addlegendentry{Intra-Inter ERC  } 
                \end{axis} 
            \end{tikzpicture}
            \caption{Performance comparison respect to length of conversation (number of utterance) on the MELD development set (variability across 10 random runs). \label{fig:length_analysis}}
            % \endminipage
        \end{figure}

        We analyzed the MELD development set to evaluate the impact of conversation length on the performance as depicted in Figure~\ref{fig:length_analysis}. Overall, our method outperforms intra- and inter-speaker methods across conversations of varying lengths. 
        It is worth noting that the performance of short dialogues (conversation length less than 15) improves significantly more than that of long dialogues.
        These results also proved the importance of ``speaker characteristic'' in short conversations lacking contextual information.
        When contextual information is limited, speaker characteristics are based on the speaker’s lexical choices, which contain explicit or implicit meaning in the sentence. An LLM can extract the speaker's characteristics by recognizing the explicit or implicit meaning conveyed in these statements.  In addition, MELD is a multiparty dataset, which contains many conversations involving more than three speakers. Based on our observations on the improvement examples, ``speaker characteristic'' is especially important in short conversations which lack much contextual information. 
    
    \subsection{Case study}
    
    Our model enhances emotion recognition accuracy, even in short conversations with limited contextual information. As show in Table~\ref{tab:improve_example},
    conversation 1041 is a short dialogue consisting of only five sentences. Our model perceives two speaker descriptions, these descriptions facilitate a more accurate identification of SPEAKER\_0's discourse, leaning towards positivity rather than anger. In addition, our architecture shows an improved capacity in predicting emotions in shorter sentences through the utilization of speaker description, particularly in cases where traditional models struggle due to the minimal information contained in expressions such as \textit{``Yeah''}, or \textit{``Okay''}.
    
        Additionally, our approach adeptly copes with scenarios where the error rate is high at the beginning of the conversation as show in Table~\ref{tab:improve_example} ($u_0$, $u_1$, $u_2$ in conversation 1061). Because contextual and speaker information is lacking at the outset of the conversation, the baseline model consistently produces incorrect initial sentences. However, with the assistance of our ``speaker description'', it delivers a better performance from the beginning of the dialogue. Experimental results prove the effectiveness and versatility of our approach across diverse conversations, including those with complex contextual information.
    
    \begin{table}[htbp]
            \small
            \centering
            \caption{
            Case study of improvement examples collected in the MELD dataset. The red and green labels refer to the incorrect and correct prediction of the models, respectively. \label{tab:improve_example} }
            \resizebox{0.99\columnwidth}{!}{%
            
                \begin{tabular}{lp{0.187\textwidth}p{0.187\textwidth}p{0.187\textwidth}lll}
                    \toprule
                   \multirow{2}{*}{\textbf{Idx}}& \multicolumn{3}{l}{\textbf{Conversation 1041}}&\multirow{2}{*}{\textbf{Label}} & \multirow{2}{*}{\textbf{BiosERC}}  & \multirow{2}{*}{\textbf{Baseline}}   \\\cline{2-4}
                   &\cellcolor{aliceblue}{\textbf{Speaker\_0}} & \multicolumn{2}{l}{\cellcolor{antiquewhite}{\textbf{Speaker\_1}} }  &&\\ \midrule
                    \texttt{d$_0$}& \multicolumn{6}{p{\columnwidth}}{
                        {  SPEAKER\_0 in the conversation comes across as someone who is \textbf{confident}, \textbf{friendly} .. to create a \textbf{relaxed atmosphere} ...}
                    }\\ \cdashline{1-7}
                   \texttt{d$_1$}& \multicolumn{6}{p{\columnwidth}}{
                        {  SPEAKER\_1 in the conversation comes across as a \textbf{friendly} .. have a strong sense of \textbf{loyalty} and \textbf{trust} in their relationships... }
                    }\\
                    \midrule 
                    \texttt{u0}& \convAA{Hey Estelle,listen}  &
                    &\texttt{neutral}& \correctpred{neutral}&\correctpred{neutral}\\
                    \texttt{u1}& & \convAB{Well! Well! Well! Joey Tribbiani! So you came back huh? }&\texttt{surprise}	 &\correctpred{surprise}	 &\wrongpred{joy} \\
                    \texttt{u2}&  \convAA{What are you talking about? I never left you! You've always been my agent!}&&  \texttt{surprise}	 &\correctpred{surprise}	 &\wrongpred{anger} \\

                     \texttt{u3}&& \convAB{Really?!}&  \texttt{surprise}	 &\correctpred{surprise}	 &\correctpred{surprise}
                    \\
                     \texttt{u4}& \convAA {Yeah! }&
                    &\texttt{joy}& \correctpred{joy}&\wrongpred{anger}\\
                    \bottomrule
                \end{tabular}
            }
            \\[0.5cm] 
            \resizebox{0.99\textwidth}{!}{%
                \begin{tabular}{lp{0.09\textwidth}p{0.112\textwidth}p{0.212\textwidth}p{0.112\textwidth}p{0.112\textwidth}lll}
                    \toprule
                   \multirow{2}{*}{\textbf{Idx}}& \multicolumn{5}{l}{\textbf{Conversation 1061}}&\multirow{2}{*}{\textbf{Label}} & \multirow{2}{*}{\textbf{BiosERC} } & \multirow{2}{*}{\textbf{Baseline} }   \\ \cline{2-6}
                   &\multicolumn{2}{l}{\cellcolor{aliceblue}{\textbf{Speaker\_0}} }   &  \multicolumn{1}{l}{\cellcolor{antiquewhite}{\textbf{Speaker\_1}}}  &  \multicolumn{2}{l}{\cellcolor{beaublue}{\textbf{Speaker\_2}}}   &&&\\
                   % &\textbf{Speaker\_0} && \textbf{Speaker\_1}  & \textbf{Speaker\_2}   &&&\\
                    \midrule
                    \texttt{d$_0$}&  \multicolumn{8}{p{1.2\columnwidth}}{
                      {  SPEAKER\_0  seems to be a very \textbf{inquisitive} and \textbf{curious person}. ... SPEAKER\_0 appears to be \textbf{quite blunt} and direct in his communication style, \textbf{not mincing words or sugarcoating} his thoughts.}
                    }\\ \cdashline{1-9}
                    \texttt{d$_1$}&  \multicolumn{8}{p{1.2\columnwidth}}{
                        {  SPEAKER\_1  seems to be a \textbf{humorous} and \textbf{light-hearted person}. ...  SPEAKER\_1 shares that they have only been with one person in their whole life, and this is met with \textbf{surprise} and \textbf{disbelief} by the other ... }
                    }\\ \cdashline{1-9}
                   \texttt{d$_2$}&   \multicolumn{8}{p{1.2\columnwidth}}{
                        {  SPEAKER\_2  seems to be a \textbf{humorous} and \textbf{light-hearted person}...  SPEAKER\_2 is someone who \textbf{enjoys having fun} and is \textbf{not afraid to poke fun} at themselves or others... }
                    }\\
                    \midrule 
                     \texttt{u0}&  \convBa{Well, what?}	&&&\texttt{neutral}& \correctpred{neutral}&\wrongpred{surprise}\\
                     \texttt{u1}&  \convBa{What?}	&&&\texttt{neutral}& \correctpred{neutral}&\wrongpred{surprise}\\
                     \texttt{u2}&\convBa{ What is it?}&&&\texttt{neutral}& \correctpred{neutral}&\wrongpred{sadness}\\
                     \texttt{u3}&\convBa{That she left you?} && &\texttt{surprise}& \correctpred{surprise}&\wrongpred{sadness}\\
                     \texttt{u4}&\convBa{That she likes women?} && &\texttt{neutral}& \wrongpred{sadness}&\wrongpred{sadness}\\[0.1cm]
                     \texttt{u5}&\convBa{That she left you for another woman that likes women?} && &\texttt{neutral}& \wrongpred{surprise}&\wrongpred{sadness}\\[0.2cm]
                     \texttt{u6}&& \convBb{Little louder, okay, I think there's a man on the twelfth floor in a coma that didn't quite hear you.}&&\texttt{anger}	 &\wrongpred{neutral}	 &\correctpred{anger} \\
                      ... & &&&\\
                    %
                    % \texttt{u7}&\convBa{Then What?}&&	&\texttt{neutral}& \correctpred{neutral}&\wrongpred{surprise}\\
                    %
                    % \texttt{u7}&&\convBb{My first time with Carol.}&&\texttt{neutral}	 &\correctpred{neutral}	 &\correctpred{neutral} \\
                    % \texttt{u8}& &&\convBc{ What?}  &\texttt{neutral}& \wrongpred{surprise}&\wrongpred{surprise}\\
                    %
                    \texttt{u11}& && \convBc{With Carol?  Oh.}	&\texttt{surprise}& \correctpred{surprise}&\wrongpred{neutral}\\
                    \texttt{u12}&  \convBa{So in your whole life, you've only been with one oh.}	&&&\texttt{surprise}& \wrongpred{neutral}&\wrongpred{neutral}\\
                    \texttt{u13}& &&\convBc{Whoah, boy, hockey was a big mistake! There was a whole bunch of stuff we could've done tonight!}	&\texttt{surprise}& \correctpred{surprise}&\wrongpred{joy}\\
                    \bottomrule
                \end{tabular}
            } 
        \end{table}

\section{Limitation\label{sec:limitation}} 
In this work, we introduce a novel method for modeling speaker characteristics based on biographical information of interlocutors in a conversation, generated by a large language model (LLM). In terms of computation time, our BiosERC method requires additional computing resources for the inference of the LLM compared to methods like \textit{Intra-inter ERC}, which utilize hidden speaker identity information. Additionally, for the scope of this paper, we have not addressed issues related to human privacy data. In realistic applications, access to conversation history should be granted and clarified by the data owner. However, we believe that, with appropriate agreements to protect users' privacy data, it is possible to obtain this permission.

\section{Conclusion\label{sec:conclusion}}
In conclusion, we proposed a novel mechanism incorporating speakers' characteristics into the ERC task, which has not been fully developed in prior research. We improve the performance of the ERC task by investigating the influence of the personality of interlocutors on emotions, considering this external knowledge as a unique feature. Our experiments on three benchmark datasets consistently yielded SOTA or competitive results, thereby substantiating the effectiveness of our proposed method. Furthermore, our model is straightforward yet highly adaptable, thus enabling its applicability to a wide range of conversation analysis tasks.

\paragraph{\textbf{Acknowledgement.}}\\  This work is supported partly by AOARD grant FA23862214039.

\bibliographystyle{IEEEtran}
\bibliography{reference}

\end{document}